\documentclass[11pt,a4paper]{article}
\usepackage[hyperref]{acl2021}
\usepackage{times}
\usepackage{latexsym}

\usepackage{pgfplots}
\usepackage{scalefnt}
\usepackage{amsfonts}
\usepackage{graphicx}
\usepackage{amsmath,bm}
\usepackage{mathrsfs}
\usepackage{multirow}
\usepackage{CJKutf8}
\usepackage{verbatim}
\usepackage{amssymb}
\usepackage{algorithm}
\usepackage{algorithmic}
\usepackage{tabu}
\usepackage{booktabs}
\usepackage{color}
\usepackage{subfigure}
\usepackage{url}

\usepackage{microtype}

\aclfinalcopy 

\title{JointGT: Graph-Text Joint Representation Learning for Text Generation from Knowledge Graphs}

\author{Pei Ke$^1$, Haozhe Ji$^1$, Yu Ran$^2$, Xin Cui$^2$, Liwei Wang$^3$, Linfeng Song$^4$, \\ \textbf{Xiaoyan Zhu}$^1$\textbf{, Minlie Huang}$^1$\thanks{\quad Corresponding author}  \\
$^1$The CoAI group, Department of Computer Science and Technology, \\
Institute for Artificial Intelligence, State Key Lab of Intelligent Technology and Systems, \\
Beijing National Research Center for Information Science and Technology, \\
Tsinghua University, Beijing 100084, China \\
$^2$Sogou Inc., Beijing, China\quad $^3$The Chinese University of Hong Kong\quad $^4$Tencent AI Lab \\
  {\tt\small \{kp17,jhz20\}@mails.tsinghua.edu.cn, \{zxy-dcs,aihuang\}@tsinghua.edu.cn} \\}

\date{}

\begin{document}
\maketitle
\begin{abstract}

Existing pre-trained models for knowledge-graph-to-text (KG-to-text) generation simply fine-tune text-to-text pre-trained models such as BART or T5 on KG-to-text datasets, which largely ignore the graph structure during encoding and lack elaborate pre-training tasks to explicitly model graph-text alignments. To tackle these problems, we propose a graph-text joint representation learning model called JointGT. During encoding, we devise a structure-aware semantic aggregation module which is plugged into each Transformer layer to preserve the graph structure. Furthermore, we propose three new pre-training tasks to explicitly enhance the graph-text alignment including respective text / graph reconstruction, and graph-text alignment in the embedding space via Optimal Transport. Experiments show that JointGT obtains new state-of-the-art performance on various KG-to-text datasets\footnote{The data, codes, and model parameters are available at \url{https://github.com/thu-coai/JointGT}.}.

\end{abstract}

\section{Introduction}

Knowledge-graph-to-text (KG-to-text) generation aims to generate high-quality texts which are consistent with input graphs \cite{gradent2017webnlg}. This task requires to simultaneously encode the graph structure and the content, and effectively leverage the input graphs in the decoding process \cite{zhao2020bridge}. As a major natural language generation (NLG) task that connects knowledge graphs and texts, this task can further promote the applicability of knowledge graphs in more realistic NLG scenarios, such as knowledge-grounded dialogue generation \cite{zhou2018ccm} and story generation \cite{guan2019seg,ji2020multihop}.

Due to the limited amount of graph-text parallel data, it's hard for typical neural text generation models to learn the alignments between source entities / relations and target tokens from scratch \cite{guo2020cyclegt,fu2020partial}. Recent work resorts to constructing general-purpose pre-trained language models for KG-to-text generation. The most common and simple way is to linearize input graphs into text sequences, and directly fine-tune text-to-text Transformer-based pre-trained models like GPT \cite{radford2018gpt,radford2019gpt2}, BART \cite{lewis2020bart} or T5 \cite{raffel2020t5} on KG-to-text datasets \cite{ribeiro2020investigate,kale2020text}. Benefiting from self-supervised pre-training on large-scale unlabelled text corpora, pre-trained language models can generate high-quality texts via simply fine-tuning, and outperform other models with sophisticated structures. 

Despite the superior performance of fine-tuning pre-trained models on KG-to-text datasets, we argue that building 
pre-trained models for KG-to-text generation still faces two major challenges: 1) \textbf{Structural information loss during encoding}. Most of the existing pre-trained models capture contextual information via bidirectional Transformers \cite{devlin2019bert}, which include full attention connections. This model structure may neglect the structural information when encoding knowledge graphs since the relation between each pair of input entities is not explicitly considered \cite{zhu2019structaware}.
2) \textbf{Absence of explicit graph-text alignments}. Existing work on pre-trained models for text generation commonly adopts auto-encoding or auto-regressive text reconstruction to learn text-text alignments, which encodes the corrupted text sequence and decodes the original sequence \cite{lewis2020bart,raffel2020t5}.
Since knowledge graphs may possess more complex structures than text sequences, it's hard to explicitly learn graph-text alignments by directly using the pre-training tasks based on text reconstruction.

Thus, we propose a graph-text joint representation learning framework called \textit{JointGT} to deal with the above challenges. 
\textbf{Firstly}, to alleviate the structural information loss during encoding, we devise a simple structure-aware semantic aggregation module at each Transformer layer to aggregate contextual information following the graph structure.
\textbf{Secondly}, we propose three pre-training tasks including graph enhanced text reconstruction, text enhanced graph reconstruction, and graph-text embedding alignment to explicitly build the connection between knowledge graphs and text sequences. The first two tasks are expected to enhance the graph-text alignment in the discrete vocabulary space, where our model is required to predict the masked information of graphs / texts based on the observed information of texts / graphs.
And the third task is designed to model the graph-text alignment in the continuous embedding space via Optimal Transport \cite{peyre2019ot} to match the hidden representations of graphs and texts. Our contributions are as follows:
\begin{itemize}
    \item We propose a novel pre-trained model called JointGT for KG-to-text generation tasks. This model adopts a structure-aware semantic aggregation module to model the structure of an input graph at each Transformer layer, and utilizes three pre-training tasks to explicitly learn graph-text alignments in the discrete and continuous spaces.
    \item We conduct experiments on the datasets of KG-to-text generation including WebNLG, WebQuestions and PathQuestions. Results show that JointGT achieves new state-of-the-art performance on KG-to-text generation.
\end{itemize}

\section{Related Work}

\noindent \textbf{KG-to-Text Generation}

\noindent 
Recent studies on KG-to-text generation tasks mainly fall into three aspects: 1) {\it Encoder modification}: To alleviate the structural information loss of sequence encoders with the input of linearized graphs
\cite{gradent2017webnlg,bayu2018gtrlstm,mory2019step}, researchers focus on 
more complex encoder structures for better graph representations, such as graph neural networks \cite{marche2018gcn,ribeiro2020globallocal} and graph Transformers \cite{rik2019agenda,schmitt2020graformer}. 2) {\it Unsupervised training}: 
researchers 
devise unsupervised training objectives to jointly learn the tasks of graph-to-text and text-to-graph conversion with non-parallel graph-text data
\cite{schmitt2020unsupervised,guo2020cyclegt,jin2020genwiki}. 3) {\it Building pre-trained models}: With the development of pre-trained NLG models such as GPT \cite{radford2018gpt,radford2019gpt2}, BART \cite{lewis2020bart} and T5 \cite{raffel2020t5}, recent work directly fine-tunes these models on graph-to-text datasets and reports impressive performance \cite{ribeiro2020investigate,kale2020text,chen2020kgpt,mager2020gpttoo}.

Compared with the existing work on pre-trained models for KG-to-text generation, our model utilizes 
pre-training methods to explicitly learn graph-text alignments instead of directly fine-tuning text-to-text pre-trained models on KG-to-text datasets.


\noindent \textbf{KG-Enhanced Pre-Trained Models}

\noindent Another line of related studies is pre-trained models enhanced by knowledge graphs for natural language understanding (NLU). The motivation of these models is to incorporate knowledge graphs into pre-trained models to facilitate the understanding of entities and relations in natural language.
Early work including ERNIE \cite{zhang2019ernie} and KnowBERT \cite{peters2019knowbert} directly uses fixed entity embeddings based on TransE \cite{bordes2013transe} or word vectors \cite{mikolov2013wordvec} during pre-training. Recent work like KEPLER \cite{wang2019kepler} and JAKET \cite{yu2020jaket} resorts to jointly pre-training graph-text representations. Specifically, they encode the textual descriptions of entities with pre-trained language models as entity embeddings and jointly optimize the knowledge embedding objective and the masked language modeling objective.

In comparison, our model focuses on joint pre-training methods on knowledge graph encoding and sequence decoding in KG-to-text generation tasks, rather than considering graph-text joint encoding methods in NLU tasks.

\section{Method}

\subsection{Task Definition and Model Overview}

Given a knowledge graph $\mathcal{G}=(\mathcal{V}, \mathcal{E})$ where $\mathcal{V}=\{e_1,e_2,\cdots,e_{|\mathcal{V}|}\}$ denotes the entity set and $\mathcal{E}=(r_{ij})_{|\mathcal{V}|\times |\mathcal{V}|}$ indicates the relations connecting the entities, and its linearized version $\mathcal{G}_{linear}=(w_1,w_2,\cdots,w_m)$ which consists of $m$ tokens, our goal is to generate a text sequence $X=(x_1,x_2,\cdots,x_n)$ which is consistent with the input graph.

\begin{figure}[!htp]
  \centering
  \includegraphics[width=1.0\linewidth]{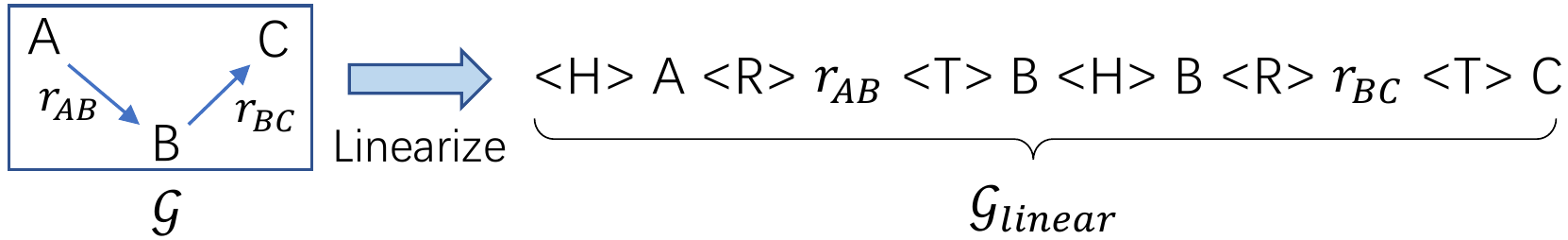}
  \caption{Illustration of linearizing knowledge graphs into text sequences. The special tokens $<$H$>$, $<$R$>$ and $<$T$>$ mean the head entity, relation and tail entity in the knowledge triples, respectively.}
  \label{fig:linearize}
\end{figure}

Our model is built on pre-trained encoder-decoder models like BART \cite{lewis2020bart} and T5 \cite{raffel2020t5}.
First of all, we follow the existing work \cite{chen2020kgpt} to linearize knowledge graphs in the form of triple lists (as shown in Figure \ref{fig:linearize}), and devise a simple structure-aware semantic aggregation module which is plugged into each Transformer layer of the encoder to preserve the structural information of input graphs (\S \ref{sec:encoder}). Then, we propose three pre-training tasks including graph / text reconstruction in the discrete vocabulary space and graph-text matching in the continuous embedding space,
which enable our model to jointly learn the representations of knowledge graphs and texts (\S \ref{sec:pretrain}).

\subsection{Model Structure}
\label{sec:encoder}

\begin{figure}[!htp]
  \centering
  \includegraphics[width=1.0\linewidth]{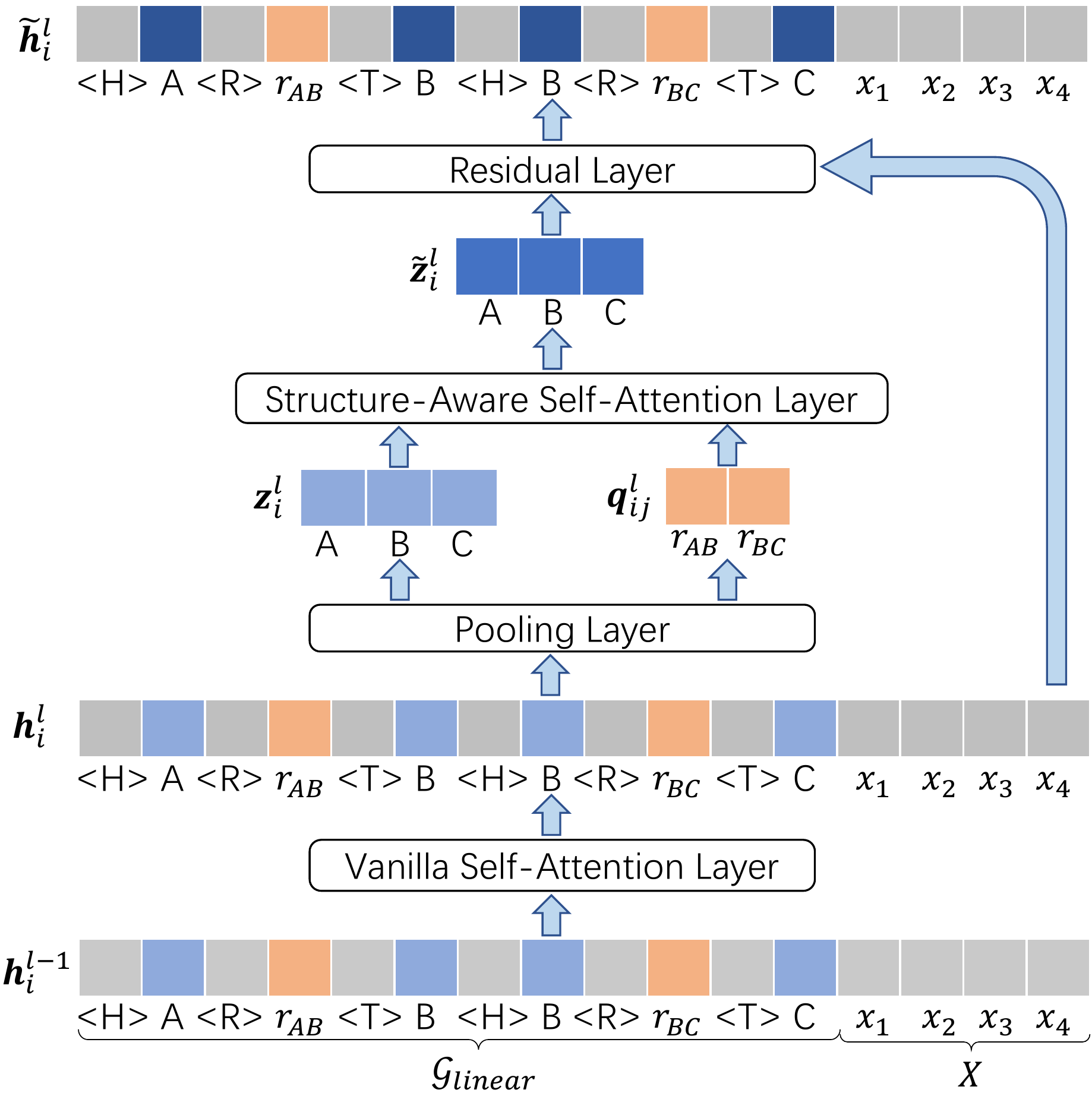}
  \caption{Structure-aware semantic aggregation module at each layer of the Transformer encoder. This module contains a pooling layer to obtain the contextual semantic representations of entities ($\bm{z}_i^l$) and relations ($\bm{q}_{ij}^l$) from the output of the vanilla self-attention layer ($\bm{h}_i^l$), a structure-aware self-attention layer to aggregate the entity representations ($\tilde{\bm{z}}_i^l$) based on the graph structure, and a residual layer to fuse the contextual and structural representations ($\tilde{\bm{h}}_i^l$).}
  \label{fig:encoder}
\end{figure}

\begin{figure*}[!htp]
  \centering
  \includegraphics[width=1.0\linewidth]{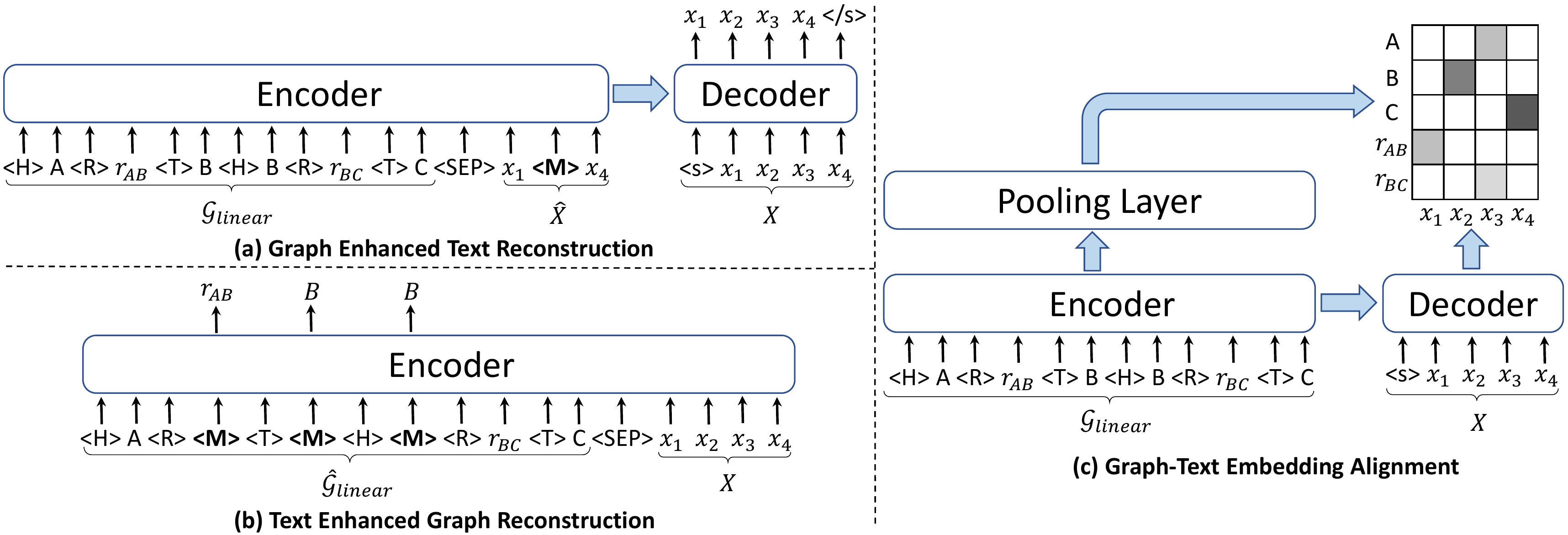}
  \caption{Overview of our proposed pre-training tasks: (a) Graph enhanced text reconstruction: reconstructing the text sequence given the complete graph. (b) Text enhanced graph reconstruction: predicting the masked entities and relations of the corrupted graph conditioned on the complete text. (c) Graph-text embedding alignment: matching the embedding vectors of the knowledge graph and the text via Optimal Transport. The special token $<$SEP$>$ is to separate the linearized graph and the text, while $<$M$>$ denotes the placeholder for masked tokens.}
  \label{fig:overview}
\end{figure*}

To simultaneously leverage the contextual representation from pre-trained models and preserve the structural information, we devise a structure-aware semantic aggregation module in the Transformer encoder. 
Assume that the input of our encoder during pre-training is the linearized graph $\mathcal{G}_{linear}$ and the corresponding text sequence $X$ (which may be corrupted or empty in some pre-training tasks), the self-attention layer in the $l$-th Transformer layer can be formulated as follows\footnote{We take a single attention head as an example in this section. In practice, we use our proposed method in the multi-head attention.}:
\begin{align}
    \bm{h}_i^l & = \sum_{j=1}^{m+n} \alpha_{ij}^l (\bm{h}_j^{l-1}\bm{W}^V) \notag \\
    \alpha_{ij}^{l} & = \frac{\exp(t_{ij}^{l})}{\sum_{p=1}^{m+n} \exp(t_{ip}^{l})} \\
    t_{ij}^{l} & = \frac{\left(\bm{h}_i^{l-1}\bm{W}^Q\right)\left(\bm{h}_j^{l-1}\bm{W}^K\right)^\top}{\sqrt{d_k}} \notag\\
    i & = 1,2,\cdots,m+n \notag
\end{align}
where $\bm{W}^Q,\bm{W}^K,\bm{W}^V$ are the model parameters
and $d_k$ denotes the dimension of query / key / value vectors. The fully-connected attention captures rich contextual semantic relationship among the entities, relations and the tokens of text sequences, but is not sufficient to encode the structural information of input graphs. Thus, we devise a structure-aware semantic aggregation module on top of vanilla self-attention, as shown in Figure \ref{fig:encoder}. First of all, we utilize a mean pooling layer\footnote{We find that there is no significant difference in the model performance between mean pooling and other aggregation functions like max pooling.} to obtain the representation of each entity and relation from the output of the vanilla self-attention layer:
\begin{align}
    \bm{z}_{i}^{l} = & \mathrm{pooling}(\{\bm{h}_p^l|p\in \mathcal{P}(e_i),1\leq p\leq m\}) \notag \\
    \bm{q}_{ij}^{l} = & \mathrm{pooling}(\{\bm{h}_p^l|p\in \mathcal{P}(r_{ij}),1\leq p\leq m\}) \notag \\
    & i=1,\cdots,|\mathcal{V}|;\quad j=1,\cdots,|\mathcal{V}|
\end{align}
%
where $\mathcal{P}(e_i) / \mathcal{P}(r_{ij})$ means the set of positions occupied by $e_i$ / $r_{ij}$ in the linearized graph.
Note that $\bm{q}_{ij}^{l}$ will be set to an all-zero vector if there is no relation between $e_i$ and $e_j$. 
Then we update entity representations with a structure-aware self-attention layer \cite{shaw2018relative}:
\begin{align}
    \tilde{\bm{z}}_{i}^l & = \sum_{j=1}^{|\mathcal{V}|}\beta_{ij}^l (\bm{z}_j^l \bm{W}^{VS} + \bm{q}_{ij}^l \bm{W}^{VR}) \notag \\
    \beta_{ij}^{l} & = \frac{\exp(u_{ij}^{l})}{\sum_{p=1}^{|\mathcal{V}|} \exp(u_{ip}^{l})}
\end{align}
\begin{align}
    u_{ij}^l & = \frac{\left(\bm{z}_i^{l}\bm{W}^{QS}\right)\left(\bm{z}_j^{l}\bm{W}^{KS}+\bm{q}_{ij}^{l}\bm{W}^{KR}\right)^\top}{\sqrt{d_k}} \notag \\
    i & = 1,2,\cdots,|\mathcal{V}| \notag
\end{align}
%
where $\bm{W}^{QS},\bm{W}^{KS},\bm{W}^{VS},\bm{W}^{KR},\bm{W}^{VR}$ are the weight matrices in the structure-aware self-attention. This layer 
integrates the contextual semantic representation of entities and relations based on the graph structure, thereby injecting the structural information into the vanilla Transformer layer. Finally, we use a residual layer to fuse semantic and structural representations of entities, and obtain the hidden states for the following computation:
\begin{eqnarray}
\tilde{\bm{h}}_{i}^l = \left\{
\begin{aligned}
\bm{h}_i^l + \tilde{\bm{z}}_j^l & , & i \in \mathcal{P}(e_j) \\
\bm{h}_i^l & , & otherwise.
\end{aligned}
\right. \\
i = 1,\cdots,m+n;\quad j=1,\cdots,|\mathcal{V}| \notag
\end{eqnarray}

Compared with existing structure-aware Transformer encoders \cite{zhu2019structaware,song2020structure} that either use the entity and relation embeddings from an external knowledge embedding model or directly learn them as model parameters, our encoder obtains the entity and relation embeddings via contextual semantic representations. This design fully employs the effective contextual representations from the existing pre-trained models while preserving the structural information, and enables our model to generalize to new entities and relations better when fine-tuned to the datasets with a different knowledge graph.


\subsection{Pre-Training Task}
\label{sec:pretrain}

Given the input graph $\mathcal{G}$ and its corresponding text sequence $X$, the goal of our pre-training task is to jointly learn the graph encoder and sequence decoder to enhance graph-text alignments, which can benefit the downstream tasks of KG-to-text generation.
We devise three pre-training tasks to explicitly learn graph-text alignments in both discrete and continuous spaces.

\subsubsection{Graph Enhanced Text Reconstruction}


The purpose of graph enhanced text reconstruction is to recover the masked text sequence based on the complete knowledge graph, as shown in Figure \ref{fig:overview}. Assume that $\hat{X}$ denotes the masked text sequence, we can formulate the loss function of this pre-training task as follows:
\begin{align}
    \mathcal{L}_{text} & = -\log P(X|\mathcal{G},\hat{X}) \notag \\
    & = -\sum_{i=1}^n \log P(x_i|\mathcal{G},\hat{X},x_{<i})
\end{align}

To construct $\hat{X}$, we masked the entity words with a probability of 40\% and other words with 20\% since entity words are more important in the task of KG-to-text generation. We also follow the existing work \cite{lewis2020bart} to merge the consecutive mask tokens into one mask token to increase the difficulty of text reconstruction. This task enables our model to utilize the knowledge graph to reconstruct the corrupted text sequence, which explores the connection between them in the discrete vocabulary space.

\subsubsection{Text Enhanced Graph Reconstruction}
%
%
As shown in Figure \ref{fig:overview}, this pre-training task aims to recover the corrupted graph according to the information of the text sequence. Given the corrupted knowledge graph $\hat{\mathcal{G}}$ with masked entities and relations, and the complete text sequence $X$, the loss function is to recover the masked entities and relations in the linearized knowledge graph:
\begin{align}
    \mathcal{L}_{graph} & = -\log P(\mathcal{G}|\hat{\mathcal{G}}, X) \notag \\
    & = -\sum_{i=1}^m M_i \log P(w_i|\hat{\mathcal{G}}, X)
\end{align}
where $M_i$ denotes an indicator function and equals 1 if and only if $w_i$ is masked. We empirically set the masking probability of entities / relations as 40\% / 20\%. This task explicitly exerts the impact of the text on the graph reconstruction, thereby guiding the encoder to focus more on the entities and relations that may appear in the text. 

\subsubsection{Graph-Text Embedding Alignment}


This pre-training task is devised to encourage the graph-text alignment in the embedding space. We use Optimal Transport (OT), which is commonly used in the cross-domain alignment \cite{chen2020got}, to calculate the minimum cost of transporting the graph representation from the encoder to the text representation from the decoder (and vice versa). As shown in Figure \ref{fig:overview}, the input of the encoder is the linearized knowledge graph $\mathcal{G}_{linear}$ while the input of the decoder is the text sequence $X$.
Assume that $\bm{H}^L=(\bm{h}_1^L,\bm{h}_2^L,\cdots,\bm{h}_{m}^L)$ indicates the final hidden states of the encoder, we can similarly acquire the entity and relation representations via mean pooling:
\begin{align}
\label{eqn:graphre}
    \bm{z}_{i}^{L}  = & \mathrm{pooling}(\{\bm{h}_p^L|p\in \mathcal{P}(e_i),1\leq p\leq m\}) \notag \\
    \bm{q}_{ij}^{L} = & \mathrm{pooling}(\{\bm{h}_p^L|p\in \mathcal{P}(r_{ij}),1\leq p\leq m\}) \notag \\
    & i=1,\cdots,|\mathcal{V}|;\quad j=1,\cdots,|\mathcal{V}|
\end{align}
%
Let $\mathcal{G}_{seq}=\mathcal{V}\cup\mathcal{E}=(g_1,g_2,\cdots,g_{|\mathcal{V}|+|\mathcal{E}|})$ denotes the sequence of all the entities and relations in $\mathcal{G}$,
we can directly obtain the contextual embedding vectors $\bm{H}^{\mathcal{G}}=(\bm{h}_1^{\mathcal{G}},\cdots,\bm{h}_{|\mathcal{V}|+|\mathcal{E}|}^{\mathcal{G}})$ for each entity and relation
from Equation \ref{eqn:graphre}.
We can also acquire the embedding vectors of $X$ from the decoder's final hidden states, which is denoted by $\bm{S}=(\bm{s}_1,\bm{s}_2,\cdots,\bm{s}_n)$.

To model the alignment between graphs and texts in the embedding space, we regard $\mathcal{G}_{seq}$ as a discrete distribution $\bm{\mu}=\sum_{i=1}^{|\mathcal{V}|+|\mathcal{E}|}\bm{a}_i\delta_{g_i}$ and $X$ as $\bm{\upsilon}=\sum_{j=1}^n \bm{b}_j\delta_{x_j}$,
%
where $\bm{a}=\{\bm{a}_i\}_{i=1}^{|\mathcal{V}|+|\mathcal{E}|}$ and $\bm{b}=\{\bm{b}_j\}_{j=1}^n$ satisfy $\sum_{i=1}^{|\mathcal{V}|+|\mathcal{E}|}\bm{a}_i=\sum_{j=1}^{n}\bm{b}_j=1$, and $\delta_{g_i}$ / $\delta_{x_j}$ indicates the Dirac function centered on $g_i$ / $x_j$. Then, we utilize the OT distance between $\bm{\mu}$ and $\bm{\upsilon}$ as the loss function, which is defined as the solution of the following problem:
%
\begin{align}
    \mathcal{L}_{OT} 
     = \min_{\bm{T}\in \Pi(\bm{a},\bm{b})} \sum_{i=1}^{|\mathcal{V}|+|\mathcal{E}|} \sum_{j=1}^n & \bm{T}_{ij}\cdot d(g_i,x_j) \\
    \Pi(\bm{a},\bm{b}) =\{\bm{T}\in \mathbb{R}_{+}^{(|\mathcal{V}|+|\mathcal{E}|) \times n} &| \bm{T}\cdot\bm{1}_{n}=\bm{a}, \notag \\
    & \bm{T}^\top\cdot\bm{1}_{|\mathcal{V}|+|\mathcal{E}|}=\bm{b}\} \notag
\end{align}
where $\bm{T}$ denotes a transport plan, $\bm{1}_{|\mathcal{V}|+|\mathcal{E}|}$ / $\bm{1}_{n}$ indicates the $(|\mathcal{V}|+|\mathcal{E}|)$ / $n$ -dimensional all-one vector respectively, and $d(g_i,x_j)$ is the cost function of transporting $g_i$ to $x_j$. We follow the existing work \cite{chen2020uniter} to adopt the cosine distance between the contextual embedding vectors of $g_i$ and $x_j$ as the cost function, which is defined as $d(g_i,x_j)=1-\frac{\bm{h}_i^\mathcal{G}\bm{s_j}}{\Vert \bm{h}_i^\mathcal{G}\Vert_2 \Vert\bm{s}_j\Vert_2}$.
%
Since the exact minimization over $\bm{T}$ is computationally intractable, we utilize IPOT algorithm \cite{xie2019ipot} to approximate the OT distance and iteratively obtain the solution of $\bm{T}$ (more details are provided in the Appendix \ref{app:ipot}). After solving $\bm{T}$, $\mathcal{L}_{OT}$ can serve as an alignment loss to optimize the model parameters. This task builds the connection between the contextual embedding vectors of knowledge graphs and texts, and explicitly promotes the graph-text alignment in the continuous space.

\begin{table*} [!h]
\centering
\scriptsize
\setlength{\tabcolsep}{1.0mm}{
\begin{tabular}{l|c|ccc|ccc|ccc|ccc}
\hline
Dataset  & \multirow{2}*{\#Param} & \multicolumn{3}{c|}{WebNLG(U)} & \multicolumn{3}{c|}{WebNLG(C)} & \multicolumn{3}{c|}{WebQuestions} & \multicolumn{3}{c}{PathQuestions} \\
\cline{1-1} \cline{3-14}
Model & & BLEU & METEOR & ROUGE & BLEU & METEOR & ROUGE & BLEU & METEOR & ROUGE & BLEU & METEOR & ROUGE \\
\hline
SOTA-NPT & - & 61.00$^\dagger$ & 42.00$^\dagger$ & 71.00$^\dagger$ & 48.00$^\dagger$ & 36.00$^\dagger$ & 65.00$^\dagger$ & 29.45$^\ddagger$ & 30.96$^\ddagger$ & 55.45$^\ddagger$ & 61.48$^\ddagger$ & 44.57$^\ddagger$ & 77.72$^\ddagger$ \\
\hline
KGPT & 177M & 64.11$^\sharp$ & 46.30$^\sharp$ & 74.57$^\sharp$ & - & - & - & - & - & - & - & - & - \\
BART & 140M & 64.55 & 46.51 & 75.13 & 56.65 & 44.51 & 70.94 & 29.61 & 31.48 & 55.42 & 63.74 & 47.23 & 77.76 \\
T5 & 220M & 64.42 & 46.58 & 74.77 & 58.66 & 46.04 & 73.06 & 28.78 & 30.55 & 55.12 & 58.95 & 44.72 & 76.58 \\
\hline
\hline
JointGT (BART) & 160M & 65.92 & 47.15 & \textbf{76.10}** & 58.55 & 45.01 & 72.31 & \textbf{30.02}* & \textbf{32.05}** & \textbf{55.60} & \textbf{65.89}** & \textbf{48.25}** & \textbf{78.87}** \\ 
JointGT (T5) & 265M & \textbf{66.14}** & \textbf{47.25}** & 75.91 & \textbf{61.01}** & \textbf{46.32}** & \textbf{73.57}** & 28.95 & 31.29 & 54.47 & 60.45 & 45.38 & 77.59 \\ 
\hline
\end{tabular}}
\caption{Results on WebNLG, WebQuestions and PathQuestions. SOTA-NPT indicates the state-of-the-art performance from the baselines without pre-training. \#Param means the number of model parameters. The results marked with $\dagger$, $\ddagger$ and $\sharp$ are re-printed from \citet{shimorina2018webnlgv2}, \citet{chen2020bignn} and \citet{chen2020kgpt}, respectively. - means that the results are not reported in the corresponding references. * indicates that our model significantly outperforms BART and T5 on the corresponding datasets (t-test, $p<0.05$), while ** means $p<0.01$.}
\label{tab:mainresult}
\end{table*}

\section{Experiment}

\subsection{Pre-training Dataset and Implementation}

We used KGTEXT \cite{chen2020kgpt} as our pre-training dataset. This dataset contains 7M graph-text data pairs, where texts are crawled from English Wikidump\footnote{\url{https://dumps.wikimedia.org}} and the corresponding knowledge graphs are acquired by querying WikiData with the Wikipedia hyperlinks of entities in the sentences. The detailed statistics of KGTEXT are shown in Table \ref{tab:datastat}.

\begin{table} [!htp]
\centering
\scriptsize
\setlength{\tabcolsep}{0.6mm}{
\begin{tabular}{lccccc}
\hline
\multirow{2}*{Dataset}  & \multirow{2}*{\#Ent} & \multirow{2}*{\#Rel} & \#Instances & \multirow{2}*{\#Triples} & \multirow{2}*{Length} \\
 & & & (Train / Valid / Test) &  &  \\
\hline
KGTEXT & 1.8M & 1,210 & 6.98M / 10K / 10K & 27.2 & 20.2 \\
\hline
WebNLG(U)  & 3,114 & 373 & 34,352 / 4,316 / 4,224 & 2.9 & 22.7 \\
WebNLG(C)  & 3,129 & 373 & 34,536 / 4,217 / 4,148 & 2.9 & 19.8 \\
WebQuestions & 25,703 & 672 & 18,989 / 2,000 / 2,000 & 5.8 & 15.0 \\
PathQuestions  & 7,250 & 378 & 9,793 / 1,000 / 1,000 & 2.7 & 14.0 \\
\hline
\end{tabular}}
\caption{Statistics of pre-training and fine-tuning datasets, including the total number of entities and relations, the data split, the average number of triples, and the average length of texts.}
\label{tab:datastat}
\end{table}

Since our model can adapt to Transformer-based pre-trained models with the encoder-decoder framework, we chose BART \cite{lewis2020bart} and T5 \cite{raffel2020t5} as the base model in this paper, which are denoted by JointGT (BART) and JointGT (T5), respectively. The hyper-parameters of the Transformer blocks were the same as BART-base and T5-base because of the limited computational resources. We initialized our model parameters with the pre-trained checkpoint of BART-base / T5-base except for the structure-aware semantic aggregation module, which was randomly initialized.
We followed BART / T5 to use Byte-Pair Encoding (BPE) vocabulary \cite{radford2019gpt2} with the size of 50,265 /  WordPiece vocabulary \cite{kudo2018sentencepiece} with the size of 32,000. The batch size was 42 / 32 for JointGT (BART) / JointGT (T5). The maximum length of linearized input graphs was 600, while the maximum length of text sequences was 64. We adopted Adam \cite{kingma2015adam} as the optimizer and set the learning rate to be 3e-5. The warmup ratio was 0.1. JointGT was pre-trained on KGTEXT for 1 epoch with the proposed pre-training tasks. It took 44 / 69 hours for JointGT (BART) / JointGT (T5) on 3 NVIDIA Quadro RTX 6000 GPUs.

\begin{table*} [!htp]
\centering
\small
\setlength{\tabcolsep}{1.0mm}{
\begin{tabular}{l|ccc|c|ccc|c}
\hline
\multirow{2}*{Model} & \multicolumn{3}{c|}{Fluency} & \multirow{2}*{$\kappa$} & \multicolumn{3}{c|}{Adequacy} & \multirow{2}*{$\kappa$} \\
\cline{2-4} \cline{6-8}
& Win (\%) & Lose (\%) & Tie (\%) & & Win (\%) & Lose (\%) & Tie (\%) & \\
\hline
JointGT (BART) vs. BART & 29.0* & 19.7 & 51.3 & 0.413 & 26.3** & 16.0 & 57.7  & 0.517  \\
JointGT (T5) vs. T5 &  23.7 & 18.7 & 57.6 & 0.405 & 22.7* & 16.3 & 61.0 & 0.424 \\
\hline
\end{tabular}}
\caption{Human evaluation on WebNLG(U). The scores indicate the percentages of win, lose and tie when JointGT is compared with other baselines. $\kappa$ is Fleiss' Kappa (all indicate moderate agreement). The scores marked with * mean $p<0.05$ while ** means $p<0.01$ in sign test.}
\label{tab:webnlghuman}
\end{table*}

\subsection{Fine-Tuning Settings}

We adopted
WebNLG, WebQuestions and Path Questions as the benchmark datasets during fine-tuning, and provided the statistics in Table \ref{tab:datastat}.

\noindent \textbf{WebNLG}: This dataset aims to convert RDF triples into a textual description. We followed the existing work \cite{chen2020kgpt} to use the version of 2.0 \cite{shimorina2018webnlgv2}. This dataset contains two official data splits: the traditional split (Unconstrained) which guarantees that there is no overlap of input graphs among train / validation / test sets,
and a more challenging split (Constrained) where the non-overlap constraint is applied to the triples of input graphs.
We denoted these two data splits as \textit{WebNLG(U)} and \textit{WebNLG(C)} in our paper. 
We followed the preprocessing steps of the existing work \cite{chen2020kgpt} to replace the underlines in the entities and relations with spaces, and split the entities and relations in a camel case into multiple words.

\noindent \textbf{WebQuestions}: This dataset \cite{yih2016webquestion,talmor2018complexwq} is the benchmark for question generation over knowledge bases (KBQG), whose purpose is to generate natural language questions about the corresponding knowledge graphs \cite{serban2016simplequestion}. 
It is constructed from two question answering datasets, i.e., WebQuestionsSP \cite{yih2016webquestion} and ComplexWebQuestions \cite{talmor2018complexwq}. These two datasets contain natural language questions, SPARQL queries and answer entities. We converted the SPARQL query to return a subgraph, and used the same preprocessing steps and data splits as the existing work \cite{kumar2019mhqg,chen2020bignn}.

\noindent \textbf{PathQuestions}: Similar to WebQuestions, the PathQuestions dataset is also the benchmark for KBQG, which is constructed from a question answering dataset \cite{zhou2018pathquestion}. The main difference is that the knowledge graph in PathQuestions is a 2-hop / 3-hop path between two entities. We used the same preprocessing steps and data splits as the existing work \cite{kumar2019mhqg,chen2020bignn}.


More detailed fine-tuning settings including the search space and the best assignment of hyper-parameters on the downstream datasets are reported in the Appendix \ref{app:hyperparam}.


\subsection{Baselines}

We chose the following two categories of models as our baselines:

\noindent \textbf{Pre-Trained Models}: We adopted KGPT \cite{chen2020kgpt}, BART \cite{lewis2020bart} and T5 \cite{raffel2020t5} as the pre-trained baselines. KGPT is a pre-trained model for KG-to-text generation, which utilizes the same pre-training dataset as our model and directly uses KG-to-text generation as the pre-training task. BART and T5, as the state-of-the-art pre-trained models for text generation, can be applied to KG-to-text generation with the input of linearized knowledge graphs and the output of text sequences \cite{ribeiro2020investigate}.

\noindent \textbf{Task-Specific Models without Pre-Training}: We also chose the state-of-the-art task-specific models without pre-training for each dataset as our baselines, including Seq2Seq with copying or delexicalisation \cite{shimorina2018webnlgv2} for WebNLG v2.0, and G2S \cite{chen2020bignn} for WebQuestions and PathQuestions.

We directly re-printed the results of baselines if they use the same datasets as ours. Otherwise, we implemented the baselines based on the codes and model parameters released by the original papers. We reported all the results of our implemented models with the mean values over 5 runs.

\subsection{Automatic Evaluation}

We followed the existing work \cite{shimorina2018webnlgv2,chen2020bignn} to use BLEU \cite{papineni2002bleu}, METEOR \cite{banerjee2005meteor} and ROUGE-L \cite{lin2004rouge} as our automatic metrics. The main results on WebNLG, WebQuestions and PathQuestions are shown in Table \ref{tab:mainresult}. We can observe that JointGT based on BART / T5 can outperform vanilla BART / T5 on most of the metrics, respectively, and obtain the state-of-the-art performance on all the datasets. This indicates that our method can promote graph-text alignments and further enhance the performance of the state-of-the-art pre-trained models on KG-to-text datasets.

\subsection{Human Evaluation}

To further evaluate the quality of generated results, we conducted human evaluation on the WebNLG(U) dataset. We followed the existing work \cite{ferreira2019comparison,ribeiro2020globallocal} to select two criteria: \textit{fluency} (whether a sentence is grammatically fluent) and \textit{adequacy} (whether a sentence clearly describes the knowledge graph). We randomly sampled 100 knowledge graphs from the test set, and collected the generated results from our models and the most competitive baseline models (i.e., BART and T5). We used the pairwise comparison between BART / T5 and JointGT (BART) / JointGT (T5). Specifically, for each pair of generated texts (one from JointGT and the other from the corresponding baseline, given the same input knowledge graph), three annotators were hired to label which text is better (i.e., win, lose or tie) in terms of the metrics mentioned above. Note that the two metrics were evaluated independently.

Results in Table \ref{tab:webnlghuman} show that JointGT can beat the corresponding baselines in both fluency and adequacy. Especially for adequacy, our model can significantly outperform BART / T5, which indicates that our model equipped with the structure-aware encoder and well-designed pre-training tasks can generate high-quality texts to describe knowledge graphs more clearly. To evaluate the agreement among different annotators, we calculated Fleiss' Kappa \cite{fleiss1971kappa} for each pairwise comparison, where the results in Table \ref{tab:webnlghuman} show moderate agreement ($0.4\leq \kappa \leq 0.6$).

\subsection{Ablation Study}

\subsubsection{Encoder Structure}

To investigate the effect of our proposed structure-aware semantic aggregation module, we fixed the pre-training tasks and compared our encoder with two Transformer-based encoders commonly used in the existing work: 

\noindent \textbf{SeqEnc}: This sequence encoder takes linearized graphs as input and ignores structural information \cite{ribeiro2020investigate,kale2020text}.

\noindent \textbf{RelEnc}: This relation-aware encoder regards the entity sequence as input and leverages the relation embedding into the self-attention layer. Both the entity and relation embedding vectors are directly learned as model parameters \cite{shaw2018relative,zhu2019structaware,song2020structure}.

\begin{table} [!htp]
\centering
\small
\setlength{\tabcolsep}{1.0mm}{
\begin{tabular}{l|c|ccc}
\hline
Model & \#Param & BLEU & METEOR & ROUGE \\
\hline
JointGT (BART) & 160M  & \textbf{65.92} & \textbf{47.15} & \textbf{76.10}  \\
\hline
 w/ SeqEnc & 140M & 64.82 & 46.87 & 75.37 \\
 w/ RelEnc & 160M & 65.17 & 47.07 & 75.69 \\
\hline
\end{tabular}}
\caption{Ablation test of different encoder structures on WebNLG(U), including our encoder, sequence encoder (SeqEnc) and relation-aware encoder (RelEnc).}
\label{tab:ablationenc}
\end{table}


Note that we only chose the encoder structures that can directly adapt to BART / T5 for fair comparison\footnote{We observed a significant performance drop if we used the encoders which are incompatible with BART / T5 (such as graph neural networks) because we had to randomly initialize the parameters of them during pre-training.}. 
Results in Table \ref{tab:ablationenc} show that our encoder structure can perform better than the other baselines. Compared with the relation-aware encoder which can also capture the structural information of knowledge graphs, our model fully utilizes the effective contextual semantic representation to initialize the entity / relation representation at each Transformer layer instead of directly using the learnable entity / relation embedding vectors. This design equips JointGT with better generalization ability during fine-tuning, thereby enhancing our performance on downstream datasets.


\begin{table} [!htp]
\centering
\small
\setlength{\tabcolsep}{1.0mm}{
\begin{tabular}{l|c|c}
\hline
\multirow{2}*{Model} & \multicolumn{2}{c}{\#Triples} \\
\cline{2-3}
 & 1-3 & 4-7 \\
\hline
JointGT (BART)  & \textbf{71.24} & \textbf{61.36}  \\
\hline
 w/ SeqEnc  & 70.83 (-0.41) & 60.11 (-1.25) \\
 w/ RelEnc  & 70.98 (-0.26) & 60.58 (-0.78) \\
\hline
\end{tabular}}
\caption{BLEU scores of three encoders on the test set of WebNLG(U) with different numbers of input triples.}
\label{tab:encodertriple}
\end{table}

To further demonstrate the effectiveness of our encoder,
we divided the test set of WebNLG(U) into two subsets according to the number of triples in knowledge graphs, and compared the performance of three encoders. Results in Table \ref{tab:encodertriple} show that the improvement margin between our encoder and other encoders is more evident when the number of input triples is large, which indicates that our model can facilitate the encoding of knowledge graphs with more complex structures.






\subsubsection{Pre-Training Task}
\label{sec:ablationtask}

\begin{table} [!h]
\centering
\small
\setlength{\tabcolsep}{1.0mm}{
\begin{tabular}{l|ccc}
\hline
Model & BLEU & METEOR & ROUGE  \\
\hline
JointGT (BART) & \textbf{65.92} & \textbf{47.15} & \textbf{76.10}  \\
\hline
\hline
w/o TextRecon & 64.22 & 46.56 & 74.96 \\
w/o GraphRecon &  65.37 & 47.09 & 75.97  \\
w/o OT & 65.03 & 47.09 & 75.83 \\ 
\hline
w/ BARTPretrain & 64.60 & 46.78 & 75.74  \\
w/ KGPTPretrain & 65.14 & 46.94 & 75.72  \\
\hline
\end{tabular}}
\caption{Ablation test of three pre-training tasks on WebNLG(U), including text / graph reconstruction and graph-text alignments via OT. BARTPretrain / KGPTPretrain means using the pre-training tasks of BART / KGPT instead of our tasks on KGTEXT. }
\label{tab:ablationtask}
\end{table}

To study the effect of three pre-training tasks, we maintained the encoder structure and removed each task respectively to test the performance. We also replaced all our pre-training tasks with the tasks of the existing work for comparison:

\noindent \textbf{BARTPretrain}: The pre-training tasks of BART including text infilling and sentence permutation \cite{lewis2020bart}. Since these tasks cannot be applied to graph data, we only used these tasks on the text data of the pre-training dataset.

\noindent \textbf{KGPTPretrain}: The pre-training task of KGPT, i.e., KG-to-text generation on the pre-training dataset \cite{chen2020kgpt}.

Results in Table \ref{tab:ablationtask} show that each of our pre-training tasks contributes to the model performance. Compared with the other two tasks, graph enhanced text reconstruction plays a more important role in the task of KG-to-text generation, which directly supervises the decoder with the conditional generation loss. We also observe an apparent performance drop if we replace our pre-training tasks with those proposed by the existing work, thereby indicating the effectiveness of our pre-training tasks to promote KG-to-text generation.

\begin{figure*}[!htp]
  \centering
  \includegraphics[width=1.0\linewidth]{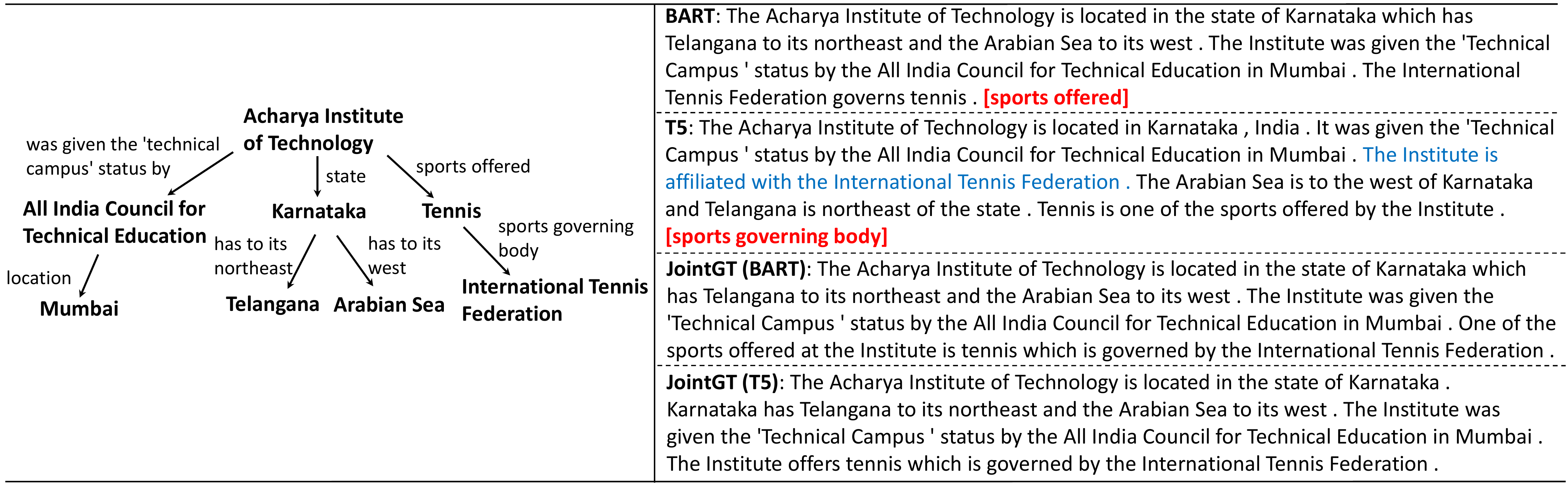}
  \caption{Generated results on WebNLG(U). We highlight the missing and unfaithful parts of each text in red and blue, respectively.}
  \label{fig:casestudy}
\end{figure*}

\subsection{Few-Shot Learning}

\begin{table} [!htp]
\centering
\small
\setlength{\tabcolsep}{1.0mm}{
\begin{tabular}{l|cccc}
\hline
\multirow{2}*{Model} & \multicolumn{4}{c}{Data Proportion} \\
\cline{2-5}
& 0.5\% & 1\% & 5\% & 10\% \\
\hline
BART & 33.92 & 39.08 & 52.24 & 56.58 \\
\hline
\hline
JointGT (BART) & \textbf{37.18} &  \textbf{42.26} & \textbf{54.41} & \textbf{57.73} \\
\hline
w/ BARTPretrain & 32.63 & 37.11 & 52.91 & 56.81 \\
w/ KGPTPretrain & 35.33 & 40.72 & 53.08 & 57.18 \\
\hline
\end{tabular}}
\caption{BLEU scores of the models with correponding pre-training tasks trained on different proportions of WebNLG(U).}
\label{tab:fewshot}
\end{table}

To further analyze whether our pre-training tasks can learn a good graph-text joint representation that benefits the downstream KG-to-text generation tasks, we considered the few-shot setting where only a few training instances were used during fine-tuning. We still fixed our model structure and compared our pre-training tasks with the tasks of BART and KGPT mentioned in \S \ref{sec:ablationtask}.

Results in Table \ref{tab:fewshot} show that our pre-training tasks can perform better than other tasks, especially when the amount of training data is small. This indicates that our proposed tasks can capture the graph-text alignments during pre-training, thereby making our model generalizable to the downstream KG-to-text datasets better with only a few training samples.

\subsection{Case Study}

To intuitively show the generation quality of our model, we provided some generated cases in Figure \ref{fig:casestudy}. We observe that JointGT can generate high-quality texts that describe the knowledge graph more completely and faithfully. For example, in the generated case on WebNLG(U), both BART and T5 fail to cover all the input triples, where BART misses the triple \textit{(Acharya Institute of Technology, sports offer, Tennis)} and T5 misses \textit{(Tennis, sports governing body, International Tennis Federation)}. Also, T5 generates non-existing facts that are unfaithful to the knowledge graph. 
Equipped with the structure-aware Transformer encoder and the well-designed pre-training tasks to learn graph-text alignments, JointGT (BART) and JointGT (T5) can generate descriptions which include all the input triples and express the relation between each pair of entities more faithfully.

\section{Conclusion}

We propose a novel graph-text joint representation learning model called JointGT for KG-to-text generation. This model plugs a simple structure-aware semantic aggregation module into the vanilla Transformer layer to preserve the structure of input graphs, and utilizes three pre-training tasks to learn graph-text alignments in the discrete vocabulary space and continuous embedding space. Experiments show that JointGT can outperform state-of-the-art pre-trained NLG models on various datasets of KG-to-text generation.

\section*{Acknowledgments}

This work was partly supported by the NSFC projects (Key project with No. 61936010 and regular project with No. 61876096). This work was also supported by the Guoqiang Institute of Tsinghua University, with Grant No. 2019GQG1 and 2020GQG0005. 

\bibliographystyle{acl_natbib}
\bibliography{acl2021}

\appendix

\section{IPOT Algorithm}
\label{app:ipot}

\begin{algorithm}[htb] 
\caption{IPOT Algorithm} 
\label{alg:ipot} 
\begin{algorithmic}[1]
\REQUIRE ~~\\
$\mathcal{G}_{seq}=\{g_i\}_{i=1}^{|\mathcal{V}|+|\mathcal{E}|}$, $X=\{x_j\}_{j=1}^n$, and their embedding vectors $\bm{H}^{\mathcal{G}}=\{\bm{h}_i^{\mathcal{G}}\}_{i=1}^{|\mathcal{V}|+|\mathcal{E}|}$, $\bm{S}=\{\bm{s}_j\}_{j=1}^n$  \\
Generalized stepsize: 1/$\beta$ \\
\STATE $\bm{\sigma}=\frac{1}{n}\bm{1}_{n}$, $\bm{T}^{(1)}=\bm{1}_{|\mathcal{V}|+|\mathcal{E}|} \bm{1}_{n}^\top$
\STATE $\bm{C}_{ij}=d(g_i,x_j)=1-\frac{\bm{h}_i^\mathcal{G}\bm{s_j}}{\Vert \bm{h}_i^\mathcal{G}\Vert_2 \Vert\bm{s}_j\Vert_2}$
\STATE $\bm{A}_{ij}=e^{-\frac{\bm{C}_{ij}}{\beta}}$
\FOR{$t = 1$ to $N$}
\STATE $\bm{Q}=\bm{A}\odot\bm{T}^{(t)}$
\FOR{$k = 1$ to $K$}
\STATE $\bm{\delta}=\frac{1}{(|\mathcal{V}|+|\mathcal{E}|)\bm{Q}\bm{\sigma}}, \bm{\sigma}=\frac{1}{n\bm{Q}^\top \bm{\delta}}$
\ENDFOR
\STATE $\bm{T}^{(t+1)}=\mathrm{diag}(\bm{\delta})\bm{Q}\mathrm{diag}(\bm{\sigma})$
\ENDFOR
\RETURN $\bm{T}$
\end{algorithmic}
\end{algorithm}

Inexact Proximal point method for Optimal Transport (IPOT) is an effective iterative method to approximate OT distance and compute the transport plan $\bm{T}$ \cite{xie2019ipot}. Given the sequence of entities and relations in the knowledge graph $\mathcal{G}_{seq}=(g_1,\cdots,g_{|\mathcal{V}|+|\mathcal{E}|})$ with its corresponding embedding vectors $\bm{H}^{\mathcal{G}}=(\bm{h}_1^{\mathcal{G}},\cdots,\bm{h}_{|\mathcal{V}|+|\mathcal{E}|}^{\mathcal{G}})$, and the text sequence $X=(x_1,\cdots,x_n)$ with its embedding vectors $\bm{S}=(\bm{s}_1,\cdots,\bm{s}_n)$, the implementation of IPOT algorithm to calculate $\bm{T}$ is shown in Algorithm \ref{alg:ipot}.

In the algorithm of IPOT, $\odot$ denotes Hadamard product. $\beta$, $K$ and $N$ are all hyper-parameters. We followed the existing work \cite{chen2020got} to set $\beta=1.0$, $K=1$ and $N=10$.

\section{Hyper-Parameter Setting}
\label{app:hyperparam}

\begin{table} [!htp]
\centering
\small
\setlength{\tabcolsep}{1.0mm}{
\begin{tabular}{cc}
\hline
Hyper-parameter & Search Space \\
\hline
Masking Probability & \multirow{2}*{\textit{choice}[20\%,30\%,40\%]} \\
(entity / relation / word) & \\
Learning Rate & \textit{choice}[2e-5,3e-5,5e-5] \\
Training Epoch & \textit{choice}[1,2] \\
Warmup Ratio & \textit{choice}[0,0.1] \\
Batch Size & \textit{choice}[32,36,42] \\
Input Length & 600 \\
Output Length & 64 \\
Maximum Gradient Norm & 1.0 \\
Optimizer & Adam \\
Epsilon (for Adam) & 1e-8 \\
\hline
\end{tabular}}
\caption{Hyper-parameter search space of JointGT during pre-training. \textit{choice} indicates that
the listed numbers will be chosen with the same probability.}
\label{tab:pretrainsearch}
\end{table}

\begin{table} [!htp]
\centering
\small
\setlength{\tabcolsep}{0.5mm}{
\begin{tabular}{cc}
\hline
Hyper-parameter & Search Space \\
\hline
Learning Rate & \textit{choice}[2e-5,3e-5,5e-5,1e-4] \\
Training Epoch & \textit{choice}[20,30,40] \\
Warmup Step & \textit{uniform-integer}[0,total\_step*0.2] \\
Batch Size & \textit{choice}[24,32] \\
Input Length & \textit{choice}[128,256] \\
Output Length & \textit{choice}[64,128] \\
Beam Size & \textit{choice}[2,3,5] \\
Length Penalty & \textit{choice}[1.0,3.0,5.0] \\
Maximum Gradient Norm & 1.0 \\
Optimizer & Adam \\
Epsilon (for Adam) & 1e-8 \\
\hline
\end{tabular}}
\caption{Hyper-parameter search space of JointGT during fine-tuning. \textit{uniform-integer} means the integers in the interval can be selected uniformly. In the search space of warmup step, total\_step denotes the total training steps on the corresponding datasets.}
\label{tab:finetunesearch}
\end{table}

We provided the detailed settings of hyper-parameters during pre-training and fine-tuning. The settings include hyper-parameter search space and best assignments. Note that we used Huggingface's Transformers\footnote{\url{https://github.com/huggingface/transformers}} to implement our models. Thus all the hyper-parameters reported in our paper were consistent with the codes of Huggingface's Transformers.

\begin{table} [!htp]
\centering
\scriptsize
\setlength{\tabcolsep}{0.5mm}{
\begin{tabular}{l|c|c|c|c}
\hline
Model & \multicolumn{4}{c}{JointGT (BART)} \\
\hline
Dataset & WebNLG(U) & WebNLG(C) & WebQuestions & PathQuestions \\
\hline
Learning Rate & 2e-5 & 2e-5 & 2e-5  & 5e-5 \\
Training Epoch & 40 & 20 & 30  & 40 \\
Warmup Step & 1,600 & 0 & 3,400 & 1,100 \\
Batch Size & 32 & 32 & 32  & 32 \\
Input Length & 256 & 256 & 256  & 128 \\
Output Length & 128 & 128 & 128  & 64 \\
Beam Size & 5 & 5 & 5  & 5 \\
Length Penalty & 1.0 & 1.0 & 5.0  & 1.0 \\
\hline
\hline
Model & \multicolumn{4}{c}{JointGT (T5)} \\
\hline
Dataset & WebNLG(U) & WebNLG(C) & WebQuestions & PathQuestions \\
\hline
Learning Rate & 5e-5 & 3e-5 & 1e-4  & 2e-5 \\
Training Epoch & 30 & 30 & 40  & 30 \\
Warmup Step & 1,600 & 1,200 & 2,300  & 900\\
Batch Size & 24 & 32 & 32  & 32 \\
Input Length & 256 & 256 & 256  & 128 \\
Output Length & 128 & 128 &  64 & 64 \\
Beam Size & 5 & 5 &  5 & 2 \\
Length Penalty & 1.0 & 1.0 & 5.0  & 1.0 \\
\hline
\end{tabular}}
\caption{Best assignments of hyper-parameters on the downstream datasets.}
\label{tab:finetunebest}
\end{table}

We presented the hyper-parameter search space during pre-training in Table \ref{tab:pretrainsearch}. The number of hyper-parameter search trials was 10. Manual search was adopted to select hyper-parameters, and the selection criterion was BLEU on the validation set when we fine-tuned the pre-trained model on WebNLG(U). The best assignment of pre-training was described in our main content.

We also provided the detailed settings of hyper-parameters during fine-tuning on the downstream datasets, including the hyper-parameter search space in Table \ref{tab:finetunesearch} and the best assignments in Table \ref{tab:finetunebest}. The number of hyper-parameter search trials was 20. BLEU was adopted as our criterion in the manual search on all the downstream tasks.

\end{document}